\documentclass[11pt]{article}

\usepackage[preprint]{acl}

\usepackage{times}
\usepackage{latexsym}

\usepackage[T1]{fontenc}

\usepackage[utf8]{inputenc}

\usepackage{microtype}

\usepackage{inconsolata}

\usepackage{graphicx}
\usepackage{multirow}
\usepackage{booktabs}
\usepackage{subcaption}
\usepackage{amsmath}

\title{Coherence Maximization Improves Pluralistic Alignment}

\author{
  Taslim Mahbub \and Yiding Pei \and Shi Feng \\
  George Washington University \\
  \texttt{\{taslim.mahbub, yidingp, shi.feng\}@gwu.edu}
}

\begin{document}
\maketitle
\begin{abstract}

Aligning AI systems with diverse human values requires value specifications grounded in concrete examples, but generating such examples without extensive human supervision remains an open challenge. We investigate what makes these examples effective, using Internal Coherence Maximization (ICM)---which infers labels by maximizing their mutual predictability---to generate persona-specific examples that steer a model toward a target group's values, without human supervision. Across four benchmarks spanning classification, preference, and open-ended generation, ICM-inferred in-context examples match the performance of gold labels. Crucially, coherence matters beyond individual label accuracy: with accuracy held constant, more coherent examples generalize substantially better than incoherent ones. 
For personas underrepresented in pretraining data, targeted human feedback on the questions where the model is least certain about a persona's values yields better generalization than the same number of labels on arbitrary questions.
These results identify coherence as a key design principle for scalable value specification, leveraging the diverse human perspectives already encoded in pretrained language models.
\end{abstract}

\section{Introduction}


\begin{figure}[t]
    \centering
    \includegraphics[width=0.47\textwidth]{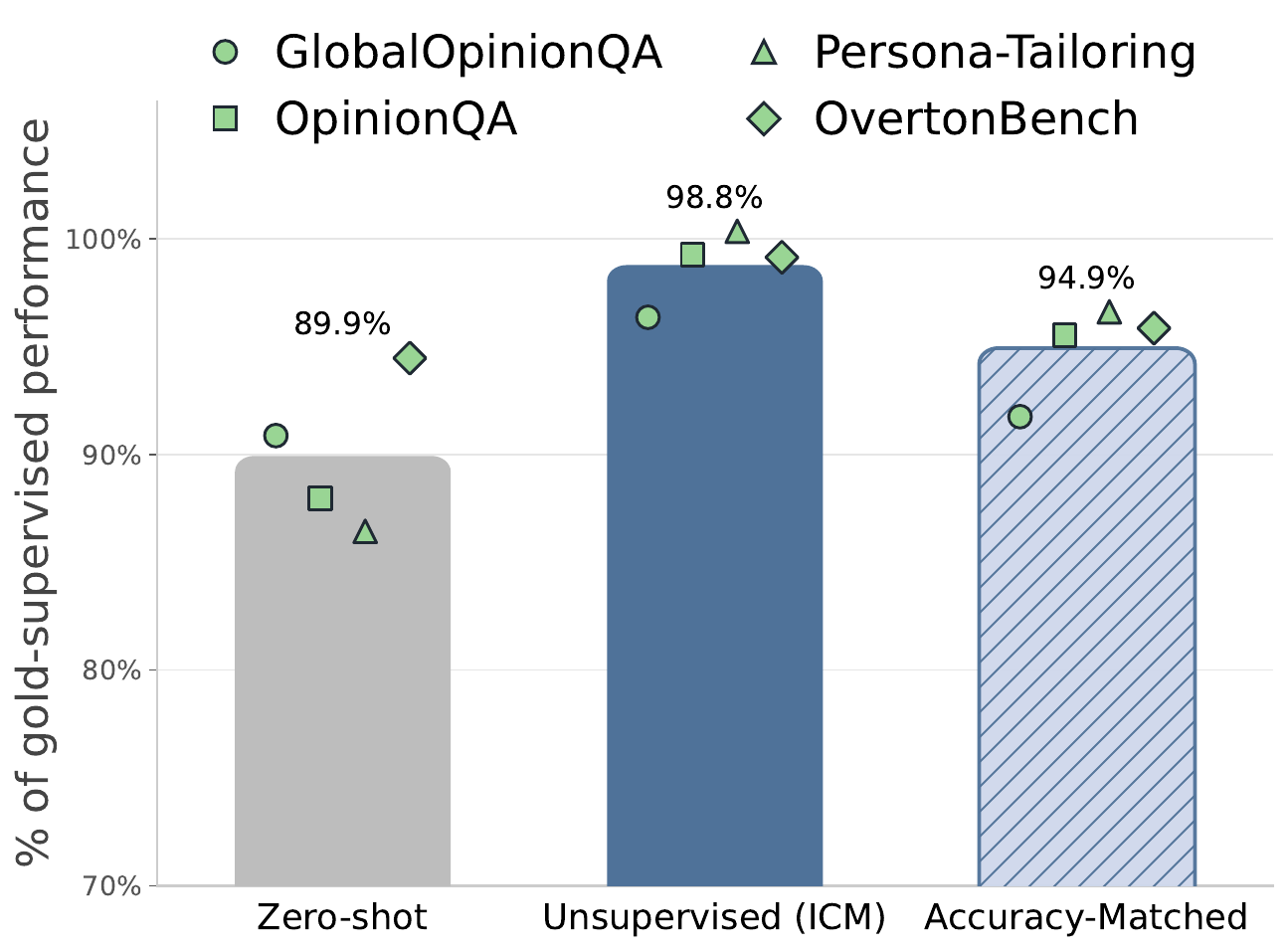}
    \caption{\textbf{Unsupervised coherence maximization matches gold-supervised performance; zero-shot prompting trails.} Each bar is the mean across all (model, dataset) pairs of a condition's score as a percentage of gold-supervised performance (y-axis starts at 70\% for visual clarity), aggregated over 6 models and 4 datasets. Our method labels the in-context examples with no human supervision, selecting the labels that are most mutually consistent under the model. The Accuracy-Matched baseline relabels the same example items to be equally accurate but less internally consistent, isolating the effect of coherence from that of raw label correctness. Markers show per-dataset means.}
    \label{fig:headline}
\end{figure}

Effective value specification remains a central challenge in AI alignment: specifications must faithfully represent human preferences while guiding models to generalize appropriately in novel contexts.
While high-level principles such as ``be helpful and harmless" provide directional guidance, recent work identifies a fundamental gap between abstract principles and practical implementation. \citet{mittelstadt2019principles} show that high-level principles lack proven methods for translation into practice, and \citet{whittlestone2019role} demonstrate that inevitable tensions arise when applying principles to concrete cases, tensions that cannot be resolved at the level of abstraction alone. This specification gap suggests that abstract principles are necessary but insufficient for effective value specification.

A growing body of evidence indicates that concrete examples are not merely supplementary but foundational to value specification. Constitutional AI requires few-shot examples to operationalize its principles \citep{bai2022constitutional}, InstructGPT's pipeline begins with supervised learning on examples before preference optimization \citep{ouyang2022training}, and recent work on inverse reinforcement learning shows that concrete examples can outperform preference data for capturing human values faithfully \citep{cheng2025inverse}. The case-based reasoning tradition in AI ethics grounds this observation theoretically, arguing that ethical evaluation requires carefully engaging with the interplay between general principles and particular cases \citep{feng2023case}. Empirical studies further show that grounding value specifications in examples provides a high-quality signal that zero-shot prompting misses \citep{adams2025steerable, yueh2025monitoring}.

The challenge intensifies in pluralistic settings where human preferences diverge across populations and contexts. A single constitution cannot capture the full range of human values, and standard alignment procedures risk collapsing the distributional pluralism necessary for serving diverse populations \citep{gabriel2020artificial, sorensen2024roadmap}. This motivates the need for value examples specific to a persona, which we define as a conditioning variable representing the values or preferences of a particular group (e.g., a country, a political affiliation, or a shared interaction style), demonstrating how abstract principles apply to that group.



\textbf{Our contribution.} We investigate the drivers of effective value specification for pluralistic alignment, using Internal Coherence Maximization (ICM) to generate persona-specific examples. We make three contributions.
\emph{First}, we show that the \emph{coherence} of an in-context example set (the mutual predictability of its labels under the base model) predicts downstream generalization better than the labels' individual \emph{accuracy}, i.e., the fraction that match ground truth: holding label accuracy fixed, more coherent example sets generalize substantially better. \emph{Second}, ICM-inferred examples, produced without any human supervision, match gold examples at steering a model toward a target group's values---measured as label or preference accuracy on the classification benchmarks and as a representation score on open-ended generation. This holds across four datasets and three task formats, remains stable across three model families and two scales each, and persists even though the inferred labels are individually less accurate than human labels. \emph{Finally}, for populations underrepresented in pretraining data, where unsupervised inference is least reliable, we show that collecting human labels on the highest-uncertainty questions generalizes substantially better than the same number of labels on arbitrary questions. Together, these results suggest that coherence is a key design principle for scalable value specification.

\section{Background and Motivation}

\paragraph{The Pluralism Problem.}
In a pluralistic world, human preferences differ across individuals, cultures, and contexts, making a single globally correct specification ill-defined \citep{conitzer2024social}. Contemporary alignment operationalizes values via global axes such as helpfulness, honesty, and harmlessness, yet these objectives generalize poorly where preferences diverge \citep{sorensen2024roadmap}. RLHF-based training compounds this: post-training datasets encode conflicting preferences \citep{movva2025s}, but standard procedures select one perspective rather than preserving diversity \citep{khan2025randomness, zhang2024diverging}.

\paragraph{Limitations of Current Approaches.}

Interactive elicitation methods such as GATE address pluralism through steerable personalization but face two limitations: they require extensive direct user interaction, and they do not leverage latent knowledge about value diversity already encoded within pretrained models. Critically, base models have been shown to better reflect human values across countries than post-trained models \citep{sorensen2024roadmap}, possibly because post-training collapses the pluralistic structure present in pretraining data. This suggests an opportunity: extracting value-relevant structure from pretrained models with minimal supervision, reserving interaction for cases where it is most needed. We hypothesize that pretrained models encode distinct value profiles for the diverse populations they encounter, and that maximizing the internal coherence of a persona's labels can surface these profiles without supervision.



\paragraph{Operationalizing Pluralism.}

We operationalize pluralistic alignment as \emph{steerability} \citep{sorensen2024roadmap}: the ability to condition a model on a target persona so that its outputs reflect that group's values rather than a single global norm. This is the property we target throughout. Concretely, steerability asks: when conditioned on a persona, does the model produce outputs consistent with that group? We measure it across all four datasets---as label or preference accuracy on the classification benchmarks and as a representation score on open-ended generation (Figure~\ref{fig:main_accuracy}). As a complementary lens, we also report \emph{distributional} pluralism, i.e., whether the model captures how opinions differ across populations, by comparing predicted and observed opinion distributions per persona (Figure~\ref{fig:persona_wise}).

\section{Method}

Our approach uses a coherence-maximization algorithm to generate persona-specific value examples without human supervision. The pipeline consists of three stages:
(1) persona extraction and item selection, (2) ICM-based label inference, and (3) in-context conditioning for inference.

\subsection{Personas and Items}

The pipeline begins by extracting coarse-grained persona features (e.g., nationality, demographics, or political affiliation) as contextual anchors. In this work, for simplicity and reproducibility, we restrict persona extraction to features that are explicitly stated in the target query.

Given a selected persona, we identify a set of \textit{items} relevant to the target query, each paired with candidate labels. An item takes one of three forms: (a) a survey question with discrete answer options (e.g., ``Should the government do more to reduce inequality?'' with answers ``Yes''/``No''); (b) a pair of candidate responses, where the label is a binary judgment of whether response A is preferred over B; or (c) a (question, candidate-answer) pair, where the label indicates whether the answer represents the persona. These items represent dimensions along which the persona may hold preferences or beliefs. In all three cases, ICM infers the label most coherent with the persona's overall value profile.

\subsection{ICM-Based Label Inference}

We apply Internal Coherence Maximization (ICM) \cite{wen2025unsupervised} to infer labels for items in an unsupervised manner using a base language model. The core intuition is that coherent value systems exhibit statistical regularity---if a persona holds position $X$ on issue $A$, this constrains their likely position on related issue $B$. ICM exploits this by searching over label assignments to maximize a coherence score: the mutual predictability of each label given all others under the base model. Starting from a random assignment, the search iteratively proposes label changes and accepts them under a simulated-annealing criterion whose temperature decays over the run; it is repeated from multiple random seeds, and the highest-scoring assignment is retained. This recovers latent structure in the model's value priors without supervised signal. We initialize the search with Llama-3.1-70B, known to work well for complex elicitation \cite{wen2025unsupervised}, hyperparameters appear in Appendix~\ref{app:icm}.

\subsection{Inference with ICM Examples}



The ICM-inferred item--label pairs are used as few-shot examples for downstream
inference. We construct a prompt containing (1) a description of the target
persona, (2) a set of example items with their ICM-inferred labels, and (3) the
target query. Depending on the task, downstream inference takes one of two modes.


\paragraph{Discrete-label prediction.} The model predicts a discrete label for a held-out item---for example, an answer option for a survey question, or an A-vs-B preference for a response pair---informed by the coherent value profile established through the in-context examples. This uses in-context learning to generalize from known item--label pairs to unseen targets; Table~\ref{tab:dataset_examples} gives examples.

\paragraph{Steered open-ended generation.} When the downstream output is free-form text rather than a discrete label, items are (prompt, candidate-answer) pairs, and ICM labels each candidate answer as representative of the persona or not. The candidate answers ICM-search selects to represent a persona serve as in-context steering examples: the model conditions on them to generate an open-ended response to the target prompt, and an LLM-as-judge scores how well that response reflects the target group. This tests whether coherent examples improve alignment when the output is free-form rather than a classification.



\section{Experimental Setup}

\subsection{Datasets and Evaluation}

We evaluate on four datasets spanning three task formats. For each we give the data source, the persona dimension, and the metric used to score it. Unless noted, we use 4-fold cross-validation stratified by persona: ICM infers labels on the training folds, which then serve as in-context examples for the held-out fold. OvertonBench instead uses a single 30/70 search/test split stratified by political group. Results are averaged across folds (where applicable) and personas.\pagebreak

\paragraph{GlobalOpinionQA (GQA).} This benchmark assesses national pluralistic alignment \cite{sorensen2024roadmap}. We abstract each country as a persona reflecting aggregated national value priors, and test whether the model can predict that persona's opinions on a given topic. We frame each instance as classification and report \emph{label-prediction accuracy}: the exact-match rate between the predicted label and the plurality label in the survey data.

\paragraph{OpinionQA (OQA).} This dataset contains questions from Pew Research's American Trends Panel, covering a wide range of political and social issues \cite{santurkar2023whose}, with political affiliation (Democrat, Republican, Independent) as the persona dimension. Affiliation is a challenging test case: it correlates with predictable positions on some issues but is internally heterogeneous on others, so group membership provides signal without deterministic prediction. We score it with the same label-prediction accuracy as GQA.

\paragraph{Persona-Tailoring (PT).} This dataset evaluates fine-grained personalization beyond coarse demographics \cite{balepur2025whose}. The task is pairwise preference: given two candidate responses, the model selects the one a persona prefers, encoded as a binary ``A preferred over B'' judgment. To ground personas in real usage rather than verbose hand-written descriptions, we cluster the original dataset by keywords into four representative persona types: Direct-Concise-Fact-first (DCF), Dialogic-Empathetic-Coach (DEC), Direct-Methodical-Step-by-step (DMS), and Dialogic-Scholarly-Nuanced (DSN). We report \emph{preference accuracy}: the rate at which the model selects the response the persona actually prefers.

\paragraph{OvertonBench (OT).} This benchmark evaluates steerable pluralistic alignment in an open-ended generation setting \cite{poole2025benchmarking}, complementing the closed-form tasks above. It covers open-ended, politically sensitive U.S.\ questions, each paired with responses from multiple LLMs, and supplies an automated LLM-as-judge whose scores correlate strongly with human ratings ($\rho=0.88$). We use political affiliation as the persona dimension; ICM operates over (question, candidate-answer) pairs, and the answers it selects as representative serve as in-context steering examples for generation at test time. We report the per-persona \emph{representation score} from the LLM-as-judge rather than the aggregate OvertonScore.

\subsection{Baselines}

We compare our ICM-based approach against the following baselines:

\begin{itemize}
    \item \textbf{Zero-shot}: The instruction-tuned chat model generates predictions conditioned on the persona, without any in-context examples.
    \item \textbf{Few-shot (ICM Labels), \emph{Ours}}: The chat model receives few-shot examples with ICM-inferred labels from the training fold.
    \item \textbf{Few-shot (Accuracy-Matched)}: The chat model receives few-shot examples whose labels match the \emph{accuracy} of the ICM labels but not their \emph{coherence}. We construct these by randomly perturbing the ICM label assignment within each persona until it reaches the same training-set accuracy ($\pm 0.5\%$) while lowering inter-label coherence, averaging results over 5 random perturbations.
    \item \textbf{Few-shot (Gold Labels)}: The chat model receives few-shot examples with ground-truth labels---defined as the majority label for each persona–question pair---for the same set of questions used in the ICM condition.
\end{itemize}
\begin{figure*}[t]
    \centering
    \includegraphics[width=0.99\textwidth]{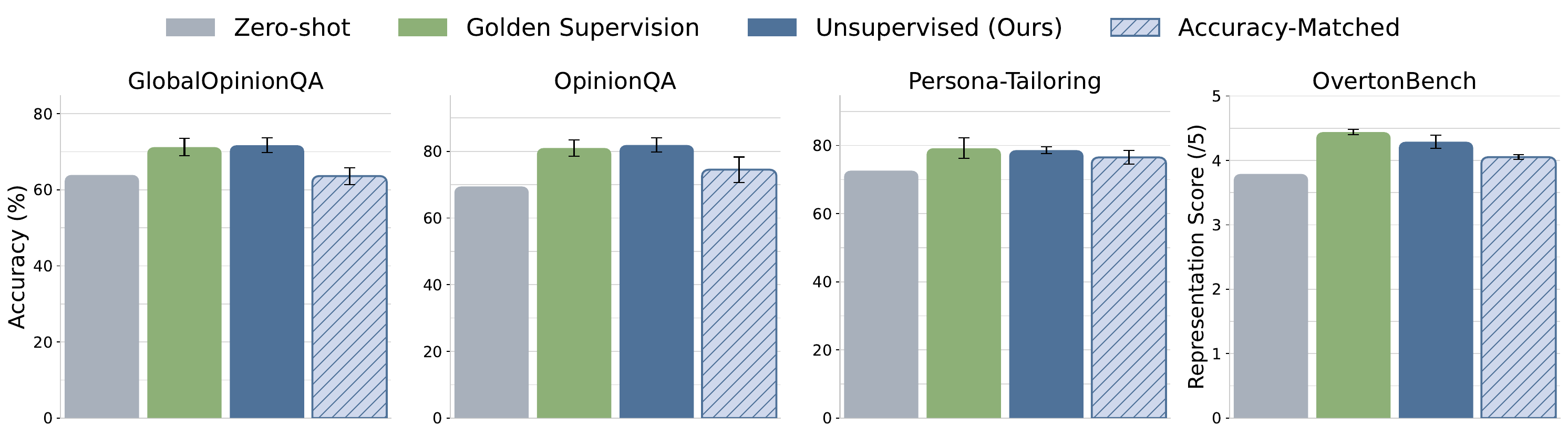}
    \caption{Test performance across four datasets and prompting conditions (Llama-3.1-70B). ICM-inferred labels (Unsupervised) match gold-supervised performance on all benchmarks and substantially outperform the zero-shot baseline (statistically significant; error bars indicate $\pm$~SE). The Accuracy-Matched condition (labels matched to ICM's accuracy but less coherent) shows that label accuracy alone does not drive performance; coherence does.}
    \label{fig:main_accuracy}
\end{figure*}

\section{Results}

Our analysis centers on two questions: (1) whether ICM matches gold-supervised performance across datasets, and (2) whether coherence, rather than label accuracy, drives that performance. An aggregated view is presented in Figure~\ref{fig:headline}.

\subsection{ICM Matches Gold Supervision}



Figure~\ref{fig:main_accuracy} presents performance across all four datasets. In the ICM, accuracy-matched, and gold conditions, the few-shot examples comprise the same items and differ only in label source. ICM-inferred labels (\emph{no human
supervision}) match gold supervision on every dataset: 71.7\% vs.\ 71.2\% on GlobalOpinionQA, 81.9\% vs.\ 81.0\% on OpinionQA, and 78.7\% vs.\ 79.2\% on Persona-Tailoring, with OvertonBench representation scores of 4.29 vs.\ 4.44.
Both conditions substantially outperform the zero-shot baseline. 

These results support our hypothesis that maximizing internal coherence surfaces the value structure implicit in each persona without explicit preference supervision. The pattern holds across all three task formats (survey classification, pairwise preference, and open-ended generation), indicating that ICM-inferred labels provide a sufficient in-context signal regardless of how the downstream task is framed. We emphasize that these are test-set generalization results, the accuracy of the ICM labels themselves against ground truth appears in Table~\ref{tab:persona_label_acc}.

A persona-level breakdown (Appendix~\ref{app:per_persona_metrics}) shows that these aggregate gains are not uniform. Most personas improve over the zero-shot baseline, but for some, elicitation does not help: in GlobalOpinionQA, Nigeria and Russia show no significant gain (and Brazil's performance drops), whereas Turkey and Jordan benefit substantially. OpinionQA shows a similar split---Democrats and Republicans gain significantly, while Independents do not.


\begin{figure*}[t]
    \centering
    \includegraphics[width=0.97\textwidth]{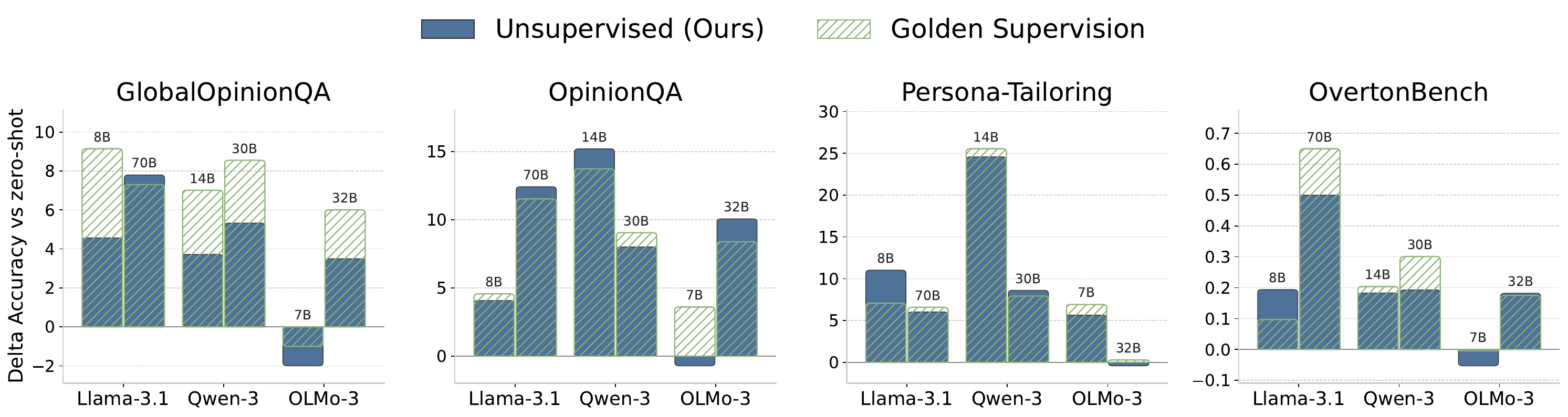}
    \caption{Improvement over the zero-shot baseline, $\Delta = \text{score} - \text{score}_{\text{zero-shot}}$, for ICM (solid) and golden supervision (hatched), overlaid per model. Each panel is a dataset; within a panel, one bar per model, grouped by family and ordered from small to large. A filled bar reaching the hatched outline means ICM matches golden supervision, exceeding it means ICM surpasses golden supervision, and a bar below zero indicates degradation relative to zero-shot. Accuracy panels (GlobalOpinionQA, OpinionQA, Persona-Tailoring) report gains in percentage points; OvertonBench reports the gain in mean representation score (0--5).}
    \label{fig:model_ablations_delta}
\end{figure*}
\subsection{Coherence, Not Accuracy, Drives Generalization}




A natural question is whether ICM's gains stem from coherence specifically or merely from having reasonably accurate labels. The Accuracy-Matched baseline isolates this: its labels match the training-set accuracy of the ICM labels ($\pm0.5\%$) but, by construction, are far less coherent.


ICM outperforms Accuracy-Matched on all four datasets despite identical label accuracy (Figure~\ref{fig:main_accuracy}). The effect is largest on the survey benchmarks---a gain of roughly eight points on both GlobalOpinionQA and OpinionQA---and smaller but directionally consistent on Persona-Tailoring and OvertonBench. Notably, on GlobalOpinionQA the accuracy-matched labels fall back to the zero-shot level, indicating that accurate but incoherent labels carry essentially no usable in-context signal. Coherence, not individual label accuracy, is the primary driver of generalization.

Beyond accuracy, coherent labels confer two further benefits, analyzed in Appendix~\ref{app:stability} and~\ref{app:sample_eff}. First, they improve \emph{prediction stability}: when each survey question's two mutually exclusive options are queried separately, ICM-conditioned models give logically consistent answers more often than gold-supervised or zero-shot conditions, while remaining well-calibrated (Figure~\ref{fig:stability}). Second, they are \emph{sample-efficient}: ICM labels surpass the zero-shot baseline with only a handful of in-context examples, and on GQA match or exceed gold labels at small budgets ($n=20, 30$; Figure~\ref{fig:n_vs_acc}).

\subsection{Robustness Across Model Families and Scales}
\label{sec:model_scaling}

Our main experiments use a single large model (Llama-3.1-70B). To check that the coherence effect is a general property rather than an artifact of one model---and that it extends to the smaller models---we repeat the full pipeline (ICM label generation and in-context inference) for three families at two sizes each: Llama-3.1 (8B, 70B), Qwen-3 (14B, 30B), and OLMo-3 (7B, 32B). Figure~\ref{fig:model_ablations_delta} reports each model's improvement over its own zero-shot baseline ($\Delta$) under both ICM and gold supervision; comparing the two shows how much of the available gain coherence maximization recovers without any labels. Absolute scores can be found in Appendix~\ref{app:model_results}.

The effect is general. Across most of the 24 model--dataset cells, ICM recovers the bulk of the gain that gold supervision provides, and it exceeds gold outright in 8 of them (Figure~\ref{fig:model_ablations_delta}). At sufficient scale, the substitution is near-complete: Llama-3.1-70B tracks gold supervision closely on every benchmark, indicating that for a capable model, unsupervised coherence maximization is a near-complete substitute for human-labeled examples.

Two boundaries emerge. GlobalOpinionQA is the hardest dataset to recover without supervision: only the largest model fully matches gold supervision, while the rest recover roughly half to two-thirds of its gain, suggesting full recovery there requires scale. And OLMo-3-7B sits below a usable capability threshold---ICM examples degrade its performance relative to zero-shot on three of four datasets---indicating the method needs a base model with enough latent value structure to surface.

Persona-Tailoring shows the opposite scale pattern: the smaller model in every family benefits more from ICM than its larger counterpart. When the task is well within a model's competence, even relatively small models can elicit useful value structure, and the advantage of scale appears only on the harder benchmarks.

\section{Collaborative Coherence Elicitation}

While the unsupervised ICM approach performs well on average, our per-persona analysis (Figure~\ref{fig:persona_wise}) reveals variation across personas. Underrepresented populations, where the base model's pretraining data may be sparse or unrepresentative, benefit least. We present an extension that incorporates minimal human feedback to improve alignment for these challenging cases.

\begin{figure*}[ht]
    \centering
    \includegraphics[width=\linewidth]{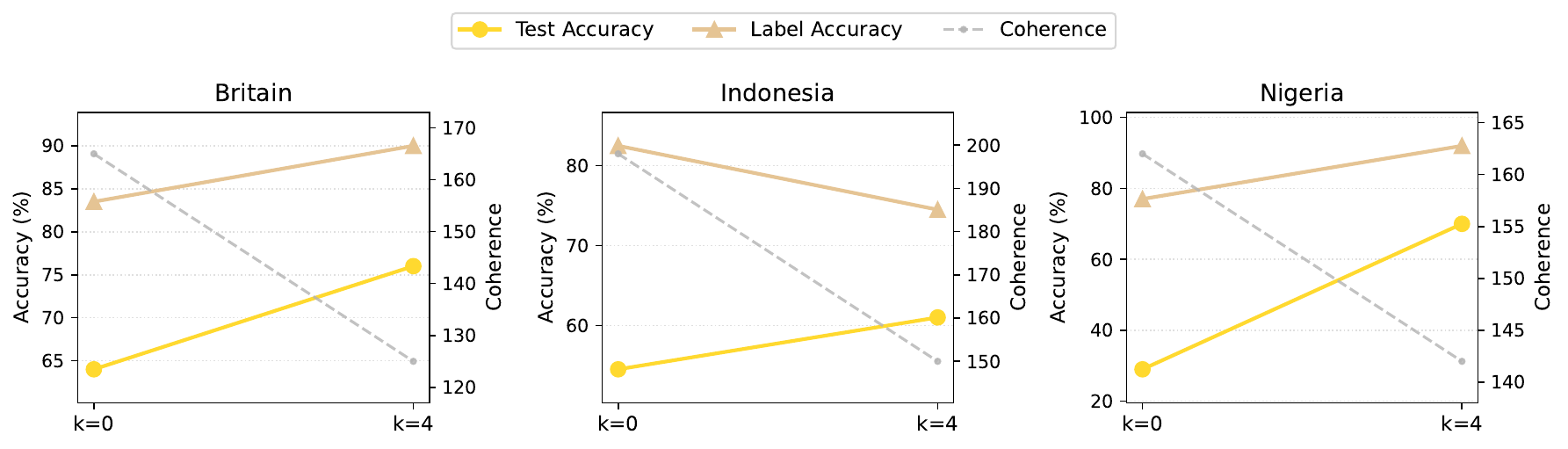}
    \caption{Effect of collaborative elicitation on the three lowest-performing personas. Test accuracy (yellow) improves substantially after incorporating 4 targeted human labels, with Nigeria showing the largest gain (+33\%). Coherence (dashed line) decreases as expected, since human-annotated labels constrain the optimization on previously uncertain dimensions. Other personas in GQA do not exhibit statistically significant changes.}
    \label{fig:gqa_interactive}
\end{figure*}

\subsection{Querying Where the Model Is Uncertain}

Interactive elicitation is an effective approach for capturing preferences on missing or ambiguous questions \cite{andukuri2024star}. However, querying users imposes a cost, and adding isolated data points may not reliably steer model behavior if they conflict with the model's existing value representation. We propose combining ICM with targeted human feedback: rather than querying users on arbitrary questions, we selectively query on \textit{low-confidence} survey questions where the model's inferred values are most uncertain.

The key insight is that ICM provides a natural confidence measure. During coherence maximization, each label is assigned based on its mutual predictability with other labels; questions where this predictability is low indicate dimensions where the model's value theory is internally inconsistent. These low-confidence questions are precisely where human input is most valuable. They represent gaps or conflicts in the model's inferred persona representation itself.

\subsection{The Elicitation Loop}

We operationalize collaborative elicitation as an iterative process between model-inferred and human-specified value judgments:

\begin{enumerate}
    \item \textbf{Initialize:} Run ICM to generate labels for all survey questions of a given persona.
    \item \textbf{Identify uncertainty:} Compute confidence for each label based on its log-probability under the mutual predictability objective. Select the $k$ lowest-confidence questions.
    \item \textbf{Query human:} Present the selected low-confidence questions to the user and collect their ground-truth labels.
    \item \textbf{Constrain and re-run:} Fix the human-provided labels as hard constraints and re-run ICM to regenerate the remaining labels, ensuring coherence with the new constraints.
\end{enumerate}

This process can be iterated; in our experiments we perform a single round, requiring only $k$ human labels per persona.

\subsection{Setup: Lowest-Performing Personas}

We evaluate collaborative elicitation on GlobalOpinionQA, focusing on the three personas with the lowest baseline ICM performance: Nigeria, Indonesia, and Britain. For each persona, we collect 1 human label per fold (4 labels total across 4-fold cross-validation), targeting the lowest-confidence question (measured using mutual predictability of individual labels) in each fold. We compare test accuracy before ($k=0$) and after ($k=4$) incorporating human-annotated data. As a baseline, we compare against the interactive pipeline with random samples and active learning (see Table~\ref{tab:interactive_performance_comparison}). For all interactive simulation experiments, we use a fixed set of hyperparameters in every search pipeline (seed=27565976, alpha=50, num-seed=8).

\subsection{Few Labels Substantially Improve Generalization}
\begin{table*}[!htbp]
\centering
\caption{Performance Comparison Across Countries and Sampling Methods.
Metrics: Test = Test-Accuracy, Label = Label-Accuracy, Coh. = Coherence.
\textbf{Bold} indicates the best value per metric within each country.}
\label{tab:interactive_performance_comparison}
\setlength{\tabcolsep}{4pt}
\begin{tabular}{l ccc ccc ccc}
\toprule
& \multicolumn{3}{c}{\textbf{Britain}}
& \multicolumn{3}{c}{\textbf{Indonesia}}
& \multicolumn{3}{c}{\textbf{Nigeria}} \\
\cmidrule(lr){2-4} \cmidrule(lr){5-7} \cmidrule(lr){8-10}
\textbf{Method} & Test & Label & Coh. & Test & Label & Coh. & Test & Label & Coh. \\
\midrule
ICM Confidence  & \textbf{0.76} & \textbf{0.90} & \textbf{125.14}
                & \textbf{0.61} & 0.75 & \textbf{150.01}
                & \textbf{0.70} & \textbf{0.92} & \textbf{142.00} \\
Active Learning & 0.72 & 0.83 & 111.54
                & 0.54 & 0.72 & 148.60
                & 0.37 & 0.61 & 107.86 \\
Random Sampling & 0.61 & 0.77 & 101.94
                & 0.54 & \textbf{0.81} & 138.19
                & 0.42 & 0.79 & 99.20 \\
\bottomrule
\end{tabular}
\end{table*}
Figure~\ref{fig:gqa_interactive} shows the results. All three personas exhibit improved test accuracy after collaborative elicitation, with Nigeria showing the most dramatic improvement, a jump of more than thirty points from a roughly even baseline, achieved with only four human labels. Indonesia improves modestly and Britain shows a substantial gain as well.

Coherence scores decrease after feedback, as expected---human labels constrain dimensions ICM previously optimized freely---but the constrained solutions are substantially more accurate, indicating that feedback corrects systematic errors in the inferred value representation. The magnitude of improvement correlates with baseline performance: Nigeria, which had the lowest initial accuracy, benefits most from human feedback.

Our interactive pipeline selects questions for human labeling based on the high uncertainty identified during ICM inference. 
We additionally compare ICM-uncertainty selection against random selection and active learning; our confidence-based selection yields the best test accuracy, label accuracy, and final coherence (Table~\ref{tab:interactive_performance_comparison}).



\section{Discussion}

\subsection{Why Coherence Outperforms Accuracy}

Our results demonstrate that coherence maximization provides a viable path toward scalable value specification for pluralistic alignment. ICM-inferred labels have lower individual accuracy than gold labels, yet they match or exceed gold-supervised performance on downstream alignment across all three task formats we study. This suggests that coherence among labels matters more than the accuracy of individual labels. This aligns with the reflective equilibrium framework motivating our approach: a value specification derives its validity not from isolated correctness on individual dimensions, but from mutual consistency across the full set of value commitments.

This pattern is robust to model scale (\S\ref{sec:model_scaling}): the coherence advantage holds across all three model families and both sizes, and ICM does not require frontier-scale models, though a minimum capability is needed to surface reliable value structure.

The success of collaborative elicitation for underrepresented personas reveals an important asymmetry: unsupervised ICM is sufficient for personas well-represented in pretraining data, but targeted human feedback becomes valuable when the model's latent priors are sparse or biased. The dramatic improvement for Nigeria suggests that small amounts of targeted feedback can correct systematic errors in the model's inferred value structure. Importantly, by querying on low-confidence questions specifically, we minimize annotation burden while maximizing impact.

\subsection{Future Work}
Several promising directions follow from this work. \textbf{Dynamic item generation} would produce relevant questions or response candidates on-the-fly rather than from a fixed pool, removing the curated-item dependence that our OvertonBench setup still assumes. \textbf{Efficient inference-time coherence} would develop lightweight approximations to the ICM search for real-time use. Finally, \textbf{individual-level personas} would extend the fine-grained interaction styles of Persona-Tailoring toward modeling distinct individual users directly.





\section{Conclusion}

We set out to investigate what properties make value specifications effective for pluralistic alignment. Using ICM to generate concrete, persona-specific examples, we demonstrated that coherent value specifications can be constructed without human annotation while matching the performance of human-annotated examples.

Across four datasets spanning survey classification, pairwise preference, and open-ended generation, ICM-inferred labels achieve alignment performance comparable to human-annotated labels. Crucially, we find that coherence among examples matters beyond individual label accuracy: when accuracy is held constant, coherent examples yield better generalization than incoherent ones.

For underrepresented personas where unsupervised inference is less reliable, incorporating targeted human feedback on high-uncertainty questions yields substantial further improvements with only a few labels per persona.

These results suggest that coherence is a key design principle for effective value specification, enabling a scalable approach to pluralistic alignment that leverages diverse human perspectives encoded in pretrained language models. By requiring no preference supervision for well-represented populations while integrating naturally with human feedback where needed, coherence maximization provides a practical pathway for building AI systems that respect the diversity of human values.




\section*{Limitations}

\paragraph{Hyperparameter Sensitivity.}
The ICM search procedure involves several hyperparameters (temperatures, number of seeds, decay rate) that may require dataset-specific tuning. While we found a single configuration that worked well across our benchmarks, generalization to substantially different domains may require re-optimization. This sensitivity could limit inference-time applications where per-dataset tuning is impractical.

\paragraph{Textual vs.\ Intentional Priors.}
Our approach extracts coherent value representations from the base model's probability distribution, which reflects statistical patterns in pretraining text rather than direct access to human intentions. While we demonstrate that these textual priors correlate with survey labels, they remain a proxy for actual human values. The pretraining corpus may encode biased or incomplete representations of global populations, particularly for regions with limited internet presence, leading to systematic underrepresentation of some majority opinions.


\section*{Acknowledgments}
We thank Tianyi Alex Qiu, Zhonghao He, and David Africa for their valuable feedback and discussions on earlier drafts of this work.


\bibliography{custom}

\appendix


\section{Related Works}
\label{sec:related}
Recent work on pluralistic alignment has pursued three complementary directions. Activation-based approaches identify linear directions encoding persona traits \citep{chen2025persona} or use Constitutional AI pipelines to shape character through DPO distillation \citep{maiya2025open}, while others address the interpretive ambiguity of such alignment rules by adapting legal statutory interpretation frameworks \citep{he2025statutory}. However, these approaches typically require explicit trait specification. Inference-time personalization methods leverage bandit algorithms over candidate pools \citep{puadurean2025inference} or psychometric activation steering \citep{zhu2024personality} to adapt to individual users without retraining. Low-supervision value extraction generates synthetic preferences from unstructured documents \citep{padhi2024value} or amplifies latent behavioral regularities through mutual information maximization \citep{franken2024self}. Evaluation frameworks like ConflictScope \citep{liu2025generative} reveal that models exhibit different value priorities in open-ended versus constrained settings, while cross-cultural studies \citep{liu2025alignment} document systematic under-alignment with global populations. The pluralism roadmap \citep{sorensen2024roadmap} formalizes steerable, overton, and distributional pluralism as distinct goals. Our coherence-based approach differs from these methods by generating persona-specific value examples through Internal Coherence Maximization (ICM)---leveraging the model's internal consistency rather than external constitutions, legal canons, documents, or psychological inventories to support pluralistic alignment without explicit preference labels.

\section{ICM Search Implementation Details}
\label{app:icm}

\paragraph{Data Processing.}
For \textsc{GlobalOpinionQA}, we select the 180 questions with complete labels across all 13 countries (Brazil, Britain, France, Germany, Indonesia, Japan, Jordan, Lebanon, Mexico, Nigeria, Pakistan, Russia, Turkey). We retain only items with \emph{binary} or \emph{three-option} answers so that each option can be mapped to a clear logical equivalence class for ICM. Five-point Likert items are excluded to avoid ambiguity: collapsing to ``Strong vs.\ Standard'' or retaining five separate classes both induce skew and yield brittle groupings.

For \textsc{OpinionQA}, we follow the same filtering pipeline and retain three major U.S.\ political affiliations (Democrat, Republican, Independent), omitting ``Other'' due to inconsistent persona definition.

After filtering, each example is converted to Alpaca-style formatting. We perform persona-stratified sampling to create 4 folds for cross-validation. For each persona–fold split, we run ICM to infer coherent labels; at test time, examples from the remaining folds (for the same persona) provide in-context examples.

\paragraph{Persona Construction for Persona-Tailoring.}
The original persona-tailoring dataset \cite{balepur2025whose} specifies personas through verbose free-text descriptions. To obtain personas grounded in recurring real-world usage patterns, we cluster these descriptions with a keyword-matching scheme: each candidate persona is defined by a set of \emph{positive} keywords (phrases indicative of the persona) and \emph{negative} keywords (phrases indicative of a contrasting persona). A dataset instance is assigned to the persona whose positive keywords it matches most strongly while matching a few negative keywords. This procedure yields four representative persona types, summarized in Table~\ref{tab:pt_keywords}.

\begin{table*}[t]
\centering
\footnotesize
\setlength{\tabcolsep}{5pt}
\caption{Keyword scheme used to cluster the Persona-Tailoring dataset
\cite{balepur2025whose} into four representative personas. Positive keywords
indicate membership; negative keywords indicate a contrasting persona.}
\label{tab:pt_keywords}
\begin{tabular}{l p{2.3cm} p{5.3cm} p{4.6cm}}
\toprule
\textbf{ID} & \textbf{Persona} & \textbf{Positive Keywords} & \textbf{Negative Keywords} \\
\midrule
DCF & Direct, Concise, Fact-first &
direct, concise, to the point, brief, short answer, no fluff, without elaboration &
explore options, background, context, nuanced, ask follow-up, conversation, discuss further \\
\addlinespace
DEC & Dialogic, Empathetic, Coach &
empathetic, gentle, reassure, acknowledge emotions, personalized, supportive, tailored &
blunt, impersonal, no empathy, strictly factual \\
\addlinespace
DMS & Direct, Methodical, Step-by-step &
step by step, step-by-step, instructions, checklist, exact measurements, clear steps, specific guidance, recipe &
high-level, theoretical, overview, open-ended, exploratory \\
\addlinespace
DSN & Dialogic, Scholarly, Nuanced &
in-depth, underlying mechanisms, nuanced, context, background, historical, clarify first, ensure understanding &
just answer, skip context, brief only, no clarifying questions \\
\bottomrule
\end{tabular}
\end{table*}

\paragraph{ICM Search Hyperparameters.}
We use the core ICM algorithm of \citet{wen2025unsupervised} with the following hyperparameters, held fixed across all models and datasets: $\alpha = 50$, $\texttt{num\_seed} = 8$, $\texttt{decay} = 0.99$, $\texttt{initial\_temp} = 10$, $\texttt{final\_temp} = 0.01$. ICM operates on the \emph{base} (pretrained, non-instruction-tuned) checkpoint of each model, producing labels from next-token log-probabilities. Our main experiments use \texttt{meta-llama/Llama-3.1-70B}; the scaling analysis
(Section~\ref{sec:model_scaling}) additionally uses five further models. All
checkpoints are listed in Table~\ref{tab:model_checkpoints}. Models are served with vLLM on NVIDIA RTX PRO 6000 GPUs; a single ICM search run takes 4--6 hours.

\begin{table*}[t]
\centering
\footnotesize
\setlength{\tabcolsep}{5pt}
\caption{Model checkpoints used for ICM label generation (base) and downstream
inference (instruction-tuned).}
\label{tab:model_checkpoints}
\begin{tabular}{llll}
\toprule
\textbf{Family} & \textbf{Size} & \textbf{Base (ICM search)} & \textbf{Instruction-tuned (inference)} \\
\midrule
\multirow{2}{*}{Llama} & 8B  & \texttt{meta-llama/Llama-3.1-8B}  & \texttt{meta-llama/Llama-3.1-8B-Instruct} \\
                       & 70B & \texttt{meta-llama/Llama-3.1-70B} & \texttt{meta-llama/Llama-3.1-70B-Instruct} \\
\midrule
\multirow{2}{*}{Qwen-3} & 14B & \texttt{Qwen/Qwen3-14B-Base}     & \texttt{Qwen/Qwen3-14B} \\
                        & 30B & \texttt{Qwen/Qwen3-30B-A3B-Base} & \texttt{Qwen/Qwen3-30B-A3B-Instruct-2507} \\
\midrule
\multirow{2}{*}{OLMo-3} & 7B  & \texttt{allenai/Olmo-3-1025-7B}  & \texttt{allenai/Olmo-3-7B-Instruct} \\
                        & 32B & \texttt{allenai/Olmo-3-1125-32B} & \texttt{allenai/Olmo-3.1-32B-Instruct} \\
\bottomrule
\end{tabular}
\end{table*}


\paragraph{Inference Model.}
For downstream evaluation we use the \emph{instruction-tuned} variant of each model (Table~\ref{tab:model_checkpoints}). This separation---base model for ICM label generation, instruction-tuned model for inference---lets us leverage the base model's preserved value diversity while benefiting from the chat model's instruction-following at test time.

\section{ICM-label Accuracy Per Persona}

Table~\ref{tab:persona_label_acc} shows the accuracy of the ICM-inferred labels that are used in-context to improve pluralistic alignment. We observe that coherent ICM-inferred examples match human-annotated performance on two benchmarks despite having lower individual accuracy on some personas.

\begin{table}[t]
\centering
\caption{ICM-inferred Label Accuracy for Llama-3.1-70B}
\begin{tabular}{|c|c|c|}
\hline
\textbf{Dataset} & \textbf{Persona} & \textbf{Label-Accuracy (\%)} \\
\hline
\multirow{13}{*}{GQA} & Brazil     & 75.00 \\
                      & Britain    & 83.50 \\
                      & France     & 84.00 \\
                      & Germany    & 88.68 \\
                      & Indonesia  & 82.35 \\
                      & Japan      & 80.81 \\
                      & Jordan     & 78.85 \\
                      & Lebanon    & 75.51 \\
                      & Mexico     & 86.54 \\
                      & Nigeria    & 76.92 \\
                      & Pakistan   & 72.45 \\
                      & Russia     & 79.38 \\
                      & Turkey     & 86.14 \\
\hline

\multirow{3}{*}{OQA} & Democrats     & 89.32 \\
                     & Republicans   & 90.12 \\
                     & Independants  & 76.47 \\
\hline
\multirow{3}{*}{PT} & DCF     & 69.11 \\
                     & DEC   & 85.75 \\
                     & DMS  & 80.75 \\
                     & DSN  & 58.38 \\
                     \hline
\multirow{3}{*}{OT} & Democrat  & 75.00 \\
                     & Independant   & 98.00 \\
                     & Republican  & 62.50 \\
\hline
\end{tabular}
\label{tab:persona_label_acc}
\end{table}


\section{Dataset Examples}
Table~\ref{tab:dataset_examples} shows 2 samples for each of the GQA (persona country = Brazil) and OQA (persona political affiliation = Republican).

\begin{table*}[t]
\centering
\caption{Sample dataset examples. GQA persona = Brazil, OQA persona = Republican, PT persona= DCF, OT persona=Democrat}
\begin{tabular}{|l|p{10cm}|}
\hline
\textbf{Dataset} & \textbf{Example} \\ \hline
GlobalOpinionQA & \textbf{Question:} Do you think women should have equal rights with men, or shouldn't they? \newline\textbf{Claim:} The majority of people from Brazil would answer this question as: 'Should' \newline I think this Claim is [True/False]. \\ \hline
OpinionQA & \textbf{Question:} Do you think men and women are basically similar or basically different when it comes to their approach to parenting? \newline\textbf{Claim:} The majority of Republicans would answer this question as: 'Men and women are basically different' \newline I think this Claim is [True/False]. \\ \hline
Persona-Tailoring & \textbf{Question:} I am trying to see how influential colours are to consumers. Do you know if any colours help consumers buy products?\newline\textbf{Human Response Preference:} Direct, Concise, Fact-first \newline \textbf{Choice 1:} Assistant: Hmm, that’s a good question. In fact, colour has a big effect on psychology. Have you heard about mood  ... \newline \textbf{Choice 2:} Assistant: I think there are two important facts here. First, humans associate colours with things that they have seen before.The second important thing to remember is that.. \newline \textbf{Claim:} Choice 1 > Choice 2. \newline I think this Claim is [True/False]. \\ \hline
OvertonBench & \textbf{Question:} In your opinion, should colleges retain their Diversity, Equity, and Inclusion (DEI) programs or eliminate them? 
\newline\textbf{Claim:} The majority of Democrat-Liberal would answer this question as: 'Colleges should maintain DEI programs to ensure campuses welcome and support students from all backgrounds. These programs ...' \newline I think this Claim is [True/False]. \\ \hline
\end{tabular}
\label{tab:dataset_examples}
\end{table*}

\section{Per-Fold Raw Accuracy Results}
Table \ref{tab:gqa_full_accuracy_results} shows the fold-wise test accuracy for GQA.

\begin{table}[ht]
\centering
\footnotesize
\setlength{\tabcolsep}{4pt}  
\caption{GQA: Accuracy (\%) Across Folds for Different Methods for Llama-3.1-70B}
\begin{tabular}{lccccr}
\hline
\textbf{Method} & \textbf{F1} & \textbf{F2} & \textbf{F3} & \textbf{F4} & \textbf{Avg} \\
\hline
Zero-shot (Base)         & 0.482 & 0.587 & 0.627 & 0.477 & 0.543 \\
Zero-shot (Chat)   & 0.603 & 0.648 & 0.640 & 0.665 & 0.639 \\
Few-shot-Rand     & 0.567 & 0.628 & 0.624 & 0.635 & 0.614 \\
Few-shot-ICM      & 0.688 & 0.730 & 0.658 & 0.791 & 0.717 \\
Few-shot ACC-ctrl & 0.646 & 0.645 & 0.658 & 0.729 & 0.670 \\
Few-shot-Gold     & 0.685 & 0.704 & 0.680 & 0.768 & 0.709 \\
\hline
\end{tabular}
\label{tab:gqa_full_accuracy_results}
\end{table}

\section{Per-Persona Performance}
\label{app:per_persona_metrics}

\begin{figure*}[t]
    \centering
    \includegraphics[width=\linewidth]{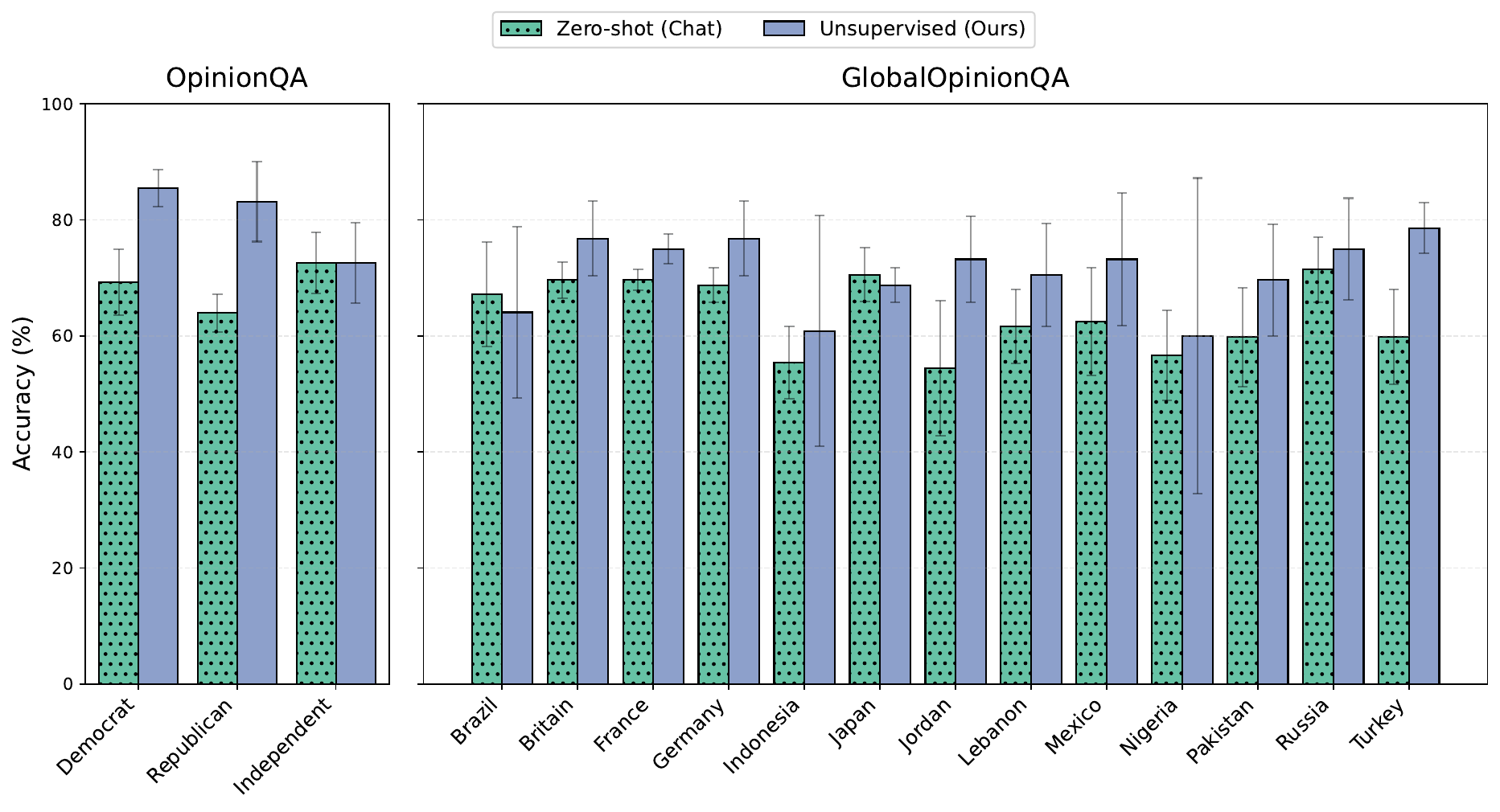}
    \caption{Test Accuracy Per Persona for GQA and OQA respectively for Llama-3.1-70B.}
    \label{fig:persona_wise}
\end{figure*}

Figure~\ref{fig:persona_wise} compares persona-wise accuracy between zero-shot inference and few-shot inference with ICM-inferred labels for GQA and OQA. We observe consistent improvements across most personas in both datasets. Notably, in GlobalOpinionsQA, the performance gains are most pronounced for personas with weaker baseline accuracy, suggesting that coherent labels provide greater benefit where the model's default priors are less reliable.

Table~\ref{tab:pt_persona_results} shows the persona-wise accuracy for zero-shot vs icm-few-shot setting in Persona-Tailoring dataset, with clear performance boost across all personas defined in our pipeline. 

\begin{table}[h]
\centering
\caption{Comparison of Zero-Shot and ICM-Few-Shot accuracy by persona on Persona-Tailoring (PT) datset for Llama-3.1-70B.}
\begin{tabular}{lcc}
\hline
\textbf{Persona} & \textbf{Zero-Shot} & \textbf{ICM} \\
\hline
DCF & 0.6718 & 0.7400 \\
DEC & 0.8500 & 0.9667 \\
DMS & 0.7875 & 0.8125 \\
DSN & 0.6579 & 0.7000 \\
\hline
\end{tabular}
\label{tab:pt_persona_results}
\end{table}

Table~\ref{tab:ot_persona_results} shows the persona-wise representation score for zero-shot prompting vs icm-few-shot setting for OvertonBench dataset. The models are generally good at inference-time steering for the Democrat Persona, but we see significant improvement for Independent and Republican personas. 

\begin{table}[h]
\centering
\caption{OvertonBench Representation Scores across Personas}
\begin{tabular}{lccc}
\toprule
Persona & Zero-Shot & ICM  \\
\midrule
Democrat    & 4.750 & 4.800  \\
Independent & 3.821 & 4.180  \\
Republican  & 2.800 & 3.880  \\
\bottomrule
\end{tabular}
\label{tab:ot_persona_results}
\end{table}

\section{Prediction Stability}
\label{app:stability}

\begin{figure}[t]
    \centering
    \includegraphics[width=\linewidth]{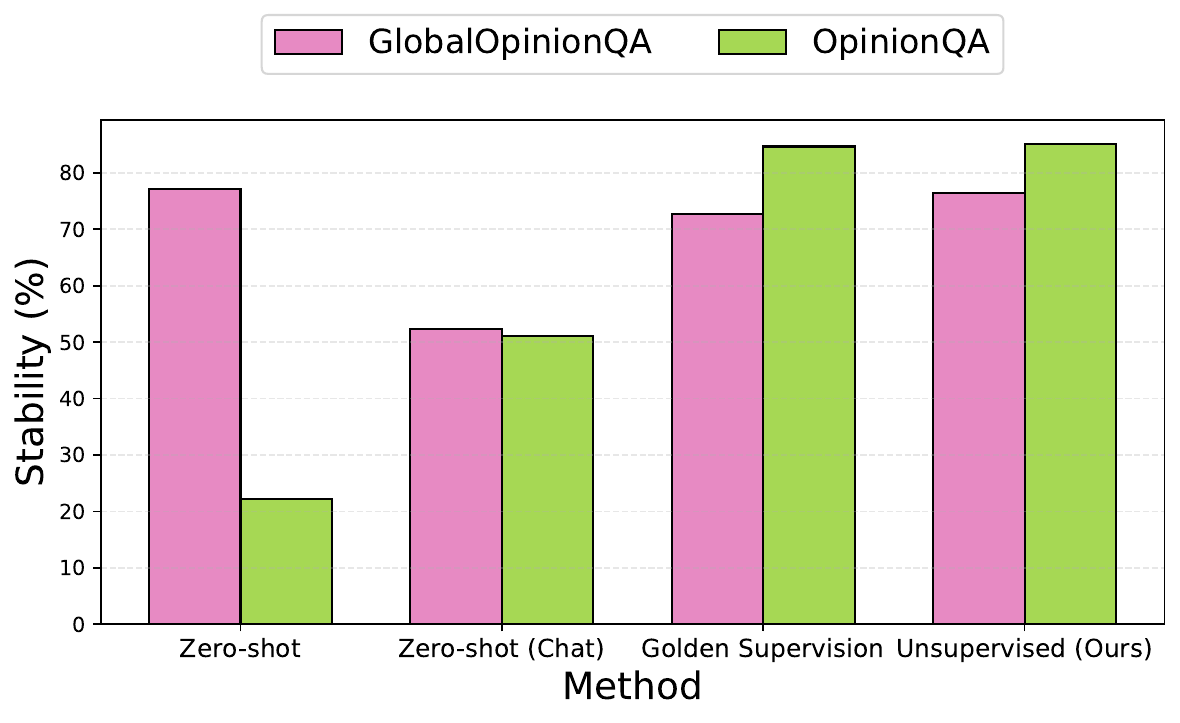}
    \caption{Coherent labels improve prediction stability. ICM-conditioned models
    produce logically consistent labels more frequently than other conditions. The Zero-shot represents the Base-model stability, and the Zero-shot (Chat) represents the instruct-model used in the regular inference pipeline.}
    \label{fig:stability}
\end{figure}

Recent work by \citet{khan2025randomness} demonstrates that LMs exhibit instability when answering cultural alignment questions. We investigate whether coherent in-context labels can mitigate this instability.

We measure stability as follows: for each survey question with two mutually exclusive options (e.g., ``Agree'' vs.\ ``Disagree''), we query the model on both options separately and check for logical consistency. A stable response labels exactly one option as True; instability occurs when the model labels both options as True or both as False.

Figure~\ref{fig:stability} shows that ICM-inferred labels produce more stable predictions than both gold-supervised labels and zero-shot baselines. Notably, for GQA the zero-shot base model exhibits the opposite pattern: high stability but low accuracy. ICM-conditioned models achieve both high stability \textit{and} high accuracy, indicating that coherent value specifications help the model maintain logical consistency without sacrificing fidelity to ground-truth preferences. We also observe lower expected calibration error (ECE) for ICM-conditioned models compared to zero-shot, with ECE comparable to gold-supervised labels (see Table~\ref{tab:ece_conditions}).

\section{Sample Efficiency of ICM-Inferred Labels}
\label{app:sample_eff}

\begin{figure}[t]
    \centering
    \includegraphics[width=\linewidth]{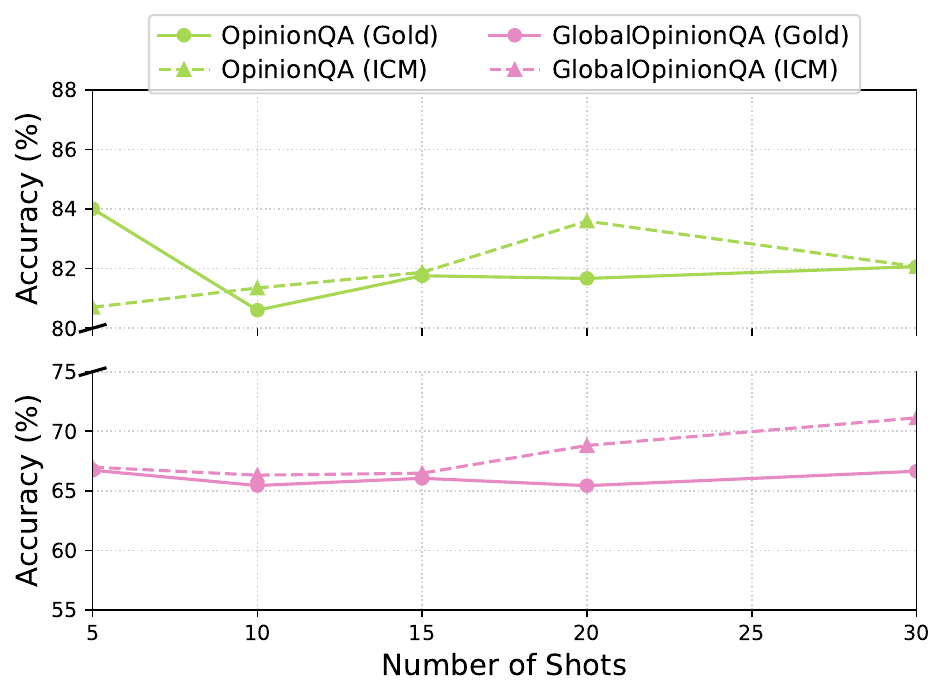}
    \caption{Test accuracy as a function of the number of in-context examples
    ($n$). GlobalOpinionQA zero-shot accuracy is 63.9\%, and OpinionQA
    zero-shot accuracy is 69.5\%. ICM labels achieve strong performance with
    relatively few examples.}
    \label{fig:n_vs_acc}
\end{figure}

We investigate how many in-context examples are needed at inference time to achieve performance gains over zero-shot baselines. Figure~\ref{fig:n_vs_acc} shows test accuracy as a function of the number of few-shot examples ($n$). On both GlobalOpinionQA and OpinionQA, ICM-inferred labels as ICL exceed zero-shot accuracy with relatively few examples. For GlobalOpinionQA, ICM labels
outperform gold labels (measured with few-shot accuracy) at $n=20$ and $n=30$, suggesting that coherent (if imperfect) labels may provide a stronger learning signal than accurate but potentially less coherent gold labels when example budgets are limited. This has practical implications: a modest number of ICM-inferred examples can meaningfully improve pluralistic alignment without requiring extensive in-context example sets.

\section{Calibration Error}

We measure Expected Calibration Error (ECE) to assess whether the model's confidence scores are well-calibrated with respect to actual accuracy. Using 15 bins, we find that ICM-conditioned models exhibit lower ECE than zero-shot, with calibration comparable to gold-supervised labels (Table~\ref{tab:ece_conditions}). This indicates that coherent in-context examples improve not only accuracy but also the reliability of the model's uncertainty estimates.

\begin{table}[!htbp]
\centering
\caption{Expected Calibration Error (ECE) across prompting conditions on two datasets for Llama-3.1-70B.}
\label{tab:ece_conditions}
\begin{tabular}{lcc}
\hline
\textbf{Condition} & \textbf{GQA ECE} & \textbf{OQA ECE} \\
\hline
Zero-shot & 0.1948 & 0.2062 \\
ICM-few        & 0.1549 & 0.1671 \\
Gold-few       & 0.1581 & 0.1457 \\
\hline
\end{tabular}
\end{table}

\section{Results Across Models and Datasets}
\label{app:model_results}

\begin{figure*}[t]
    \centering
    \includegraphics[width=0.99\textwidth]{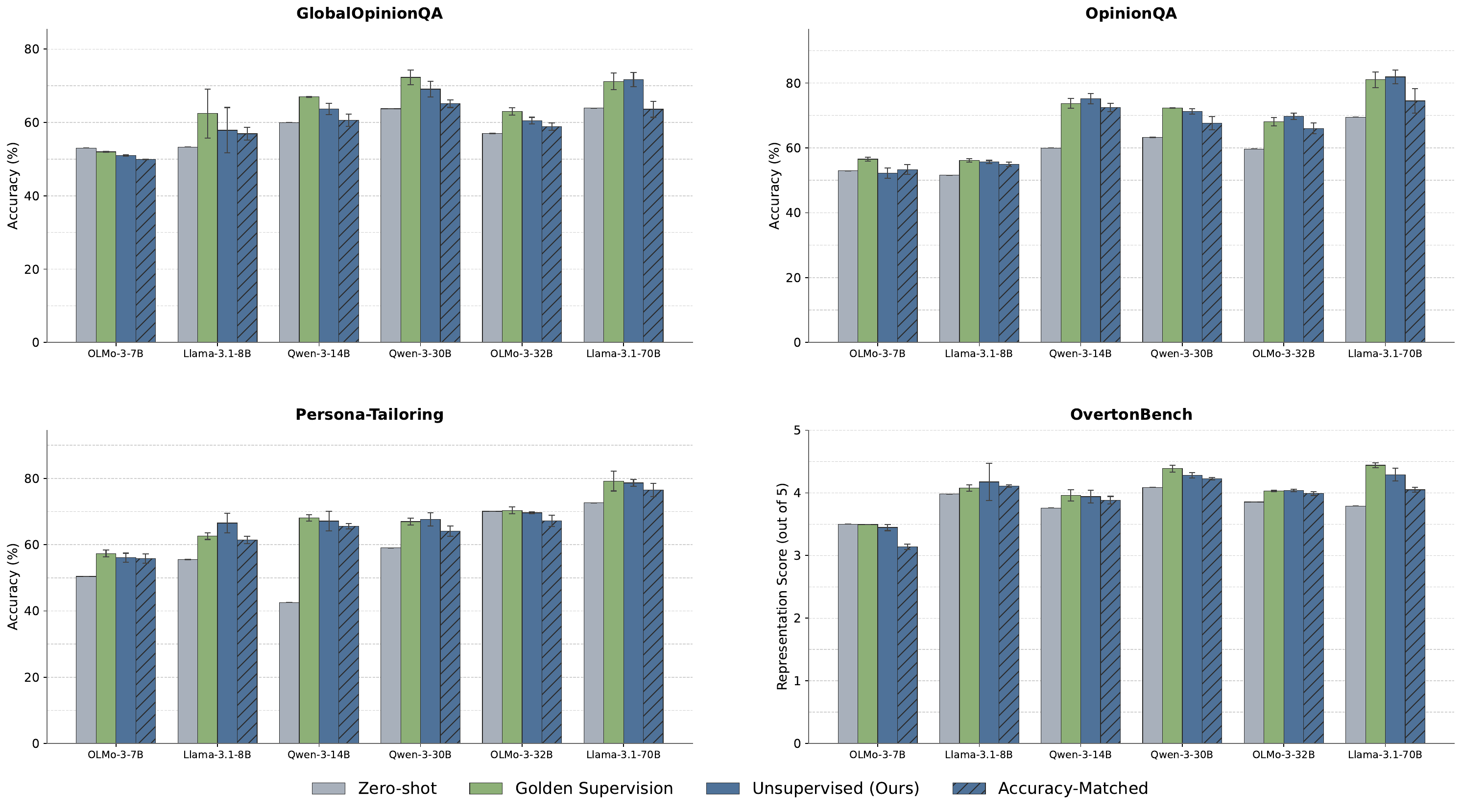}
    \caption{Absolute performance across six models and four datasets for all
    prompting conditions. GQA/OQA/PT report accuracy (\%); OvertonBench reports
    representation score (0--5).}
    \label{fig:model_ablations_all_scores}
\end{figure*}

Figure~\ref{fig:model_ablations_all_scores} reports absolute performance for all six models, four datasets, and four prompting conditions.

A few details support the claims in Section~\ref{sec:model_scaling}. ICM exceeds gold supervision in several model--dataset cells---e.g., Llama-3.1-70B and OLMo-3-32B on OpinionQA, and Llama-3.1-8B and Qwen-3-30B on Persona-Tailoring---indicating that coherent labels can occasionally provide a stronger in-context signal than human annotation. The Accuracy-Matched condition is lower than ICM in 23 of 24 cells; the sole exception is OLMo-3-7B on OpinionQA, where all conditions sit near chance. OLMo-3-7B is the weakest model throughout, performing at or below its zero-shot baseline on GlobalOpinionQA and OvertonBench, suggesting that ICM requires a base model with sufficient latent value structure to surface.

\section{Question Selection Strategies for Collaborative Elicitation}
\label{app:selection_comparison}

Our interactive pipeline selects questions for human labeling based on the high uncertainty identified during ICM inference. We compare this to two alternatives: random question selection, and active learning, which selects questions where the few-shot model has low prediction confidence. 


Table~\ref{tab:interactive_performance_comparison}
reports results for the three personas where differences were statistically significant. Confidence-based selection outperforms both baselines on test accuracy, label accuracy, and final label coherence.





\end{document}